\documentclass[letterpaper, 10 pt, conference]{IEEEtran}  

\usepackage{amsmath,amsfonts}
\usepackage{algorithmic}
\usepackage{algorithm}
\usepackage{array}
\usepackage[caption=false,font=normalsize,labelfont=sf,textfont=sf]{subfig}
\usepackage{textcomp}
\usepackage{stfloats}
\usepackage{bm}
\usepackage{url}
\usepackage{graphicx}
\usepackage{textgreek}
\usepackage{cite}
\usepackage{xspace}
\usepackage{hyperref}
\usepackage{acro}
\usepackage{siunitx}

\DeclareAcronym{hri}{
  short=HRI,
  long=Human-Robot Interaction,
}
\DeclareAcronym{mis}{
  short=MIS,
  long=minimally Invasive Surgery,
}
\DeclareAcronym{ramis}{
  short=RAMIS,
  long=RAMIS,
}
\DeclareAcronym{vf}{
  short=VF,
  long=Virtual Fixture,
}
\DeclareAcronym{ha}{
  short=HA,
  long=Haptic Assistance,
}
\DeclareAcronym{psm}{
  short=PSM,
  long=Patient-Side Manipulator,
}
\DeclareAcronym{mtm}{
  short=MTM,
  long=Master Tool Manipulator,
}
\DeclareAcronym{hrsv}{
  short=HRSV,
  long=High-Resolution Stereo Viewer,
}
\DeclareAcronym{ecm}{
  short=ECM,
  long=Endoscope Camera Manipulator,
}
\DeclareAcronym{ee}{
  short=EE,
  long=End-Effector,
}

\newcommand{\cright}{\textsuperscript{\textregistered}\xspace}
\newcommand{\vect}[1]{\textbf{\textit{#1}}}

\newcommand{\ha}{\ac{ha}\xspace}

\title{\LARGE \bf Implementation and Assessment of an Augmented Training Curriculum for Surgical Robotics}

\newcommand{\ee}{\ac{ee} \xspace}

\begin{document}
\author{Alberto Rota, Ke Fan, Elena De Momi
\thanks{\\Alberto Rota, Ke Fan and Elena De Momi are with the Department of Electrionics, Information and Bioengineering at Politecnico di Milano, Milan, Italy
\\
\\Correspondence: \texttt{alberto1.rota@polimi.it}}}
\maketitle
\thispagestyle{empty}
\pagestyle{empty}
\begin{abstract}
The integration of high-level assistance algorithms in surgical robotics training curricula may be beneficial in establishing a more comprehensive and robust skillset for aspiring surgeons, improving their clinical performance as a consequence. This work presents the development and validation of a haptic-enhanced Virtual Reality simulator for surgical robotics training, featuring 8 surgical tasks that the trainee can interact with thanks to the embedded physics engine. This virtual simulated environment is augmented by the introduction of high-level haptic interfaces for robotic assistance that aim at re-directing the motion of the trainee's hands and wrists toward targets or away from obstacles, and providing a quantitative performance score after the execution of each training exercise.  

An experimental study shows that the introduction of enhanced robotic assistance into a surgical robotics training curriculum improves performance during the training process and, crucially, promotes the transfer of the acquired skills to an unassisted surgical scenario, like the clinical one.
\end{abstract}

\begin{IEEEkeywords}
Surgical Robotics, Enhanced Training, Haptic Assistance, Skill Transfer
\end{IEEEkeywords}

\section{Introduction}
\IEEEPARstart{T}{he}
increase of surgical robotics procedures in the last decade demands a high number of trained surgeons \cite{Dupont2021} \cite{Wolfe1993}, capable of teleoperating such advanced and complex systems and at the same time able to take advantage of the benefits of Robot-Assisted Minimally Invasive Surgery (RAMIS) safely and effectively. \cite{Burke2021}


This work studies and evaluates the role of \ha strategies \cite{Rosenberg1993}\cite{Bowyer2013}, in the context of enhancing surgical robotics training for providing an increased retention and transfer of surgical skills. The haptic interface consists of high-level algorithms that assist the surgeon by providing haptic guidance at the level of the \textit{leader} manipulator, generating mechanical forces and torques which re-direct the motion of the surgeon's hands. \ha may be most beneficial in the training process that aspiring surgeons undertake \cite{Hong2016}, which often takes place in a simulated virtual environment \cite{hamza2019survey}. Exploiting the customizability of simulated surgical tasks together with the implementation of an augmented training protocol will enhance the process of learning key surgical skills \cite{McKnight2020}, in terms of performance \cite{van2009value}, retention and transfer \cite{Haque2006}. 

\section{State Of The Art}
Surgical robotics training programs have historically been conducted on animal models: Hanly \textit{et al.} \cite{hanly2004multiservice} demonstrated their validity in the context of learning key surgical robotics skills, while Patel \textit{et al.} \cite{patel2009robotic} discussed the role of animal model training for refining the surgical skillset prior to the human application.
The use of animals allows to train advanced surgical skills like dissection and homeostasis but are limited in terms of anesthesia time and anatomical similarities \cite{hoznek2003laparoscopic}. 

Human cadavers represent the most valid solution for these limitations, as they provide the highest level of realism and the possibility of performing complete surgical procedures: Blaschko \textit{et al.} found that the coordinated, multiple-use of human anatomic material represents a very effective training format \cite{blaschko2007coordinated}.
More recently, the use of cadavers in training urological robotic surgeon was correlated to  immediate performance improvements on a one-day training session \cite{bertolo2018single}.

Both the approaches with animal models and cadavers are cost-ineffective, non-reusable and non-portable: surgical training have so evolved into employing simplified but highly practical dry-lab phantoms (also known as ``laparoscopic box trainers'' or ``pelvis trainers''), which sacrifice anatomical resemblance in favor of reusability, reproducibility and costs \cite{nigicser2016anatomically}, and at the same time encouraging concentration \cite{sandor2010minimally}.
The training on a dry-lab laparoscopic pelvic trainer was demonstrated to improve the surgical skills of residents, with beneficial effects on skill retention as well \cite{dubuisson2016laparoscopic}. Inanimate training phantoms can be augmented with the integration of sensors and visual feedback systems (LED lights or similar), which contribute to the skill learning and skill transfer paradigms \cite{caccianiga2020evaluation}. Moreover, novel trends in manufacturing technology like 3D printing are making dry-lab phantoms more customizable, affordable and targetable to specific clinical training requirements \cite{smith20203d}. Inanimate training setups may cover an important role in robustly establishing the basic skills:  however, these setups are not complex enough for building an extensive and complete skillset, and lack the possibility of performing simulations of complete surgeries.   

A compromise between high realism, reproducibility and customizability is found in virtual reality (VR) simulations: crucially, VR is the only approach that allows for performance quantification, a key factor in assessing the training progress and monitoring the learning curve \cite{fu2023recent}\cite{ruikar2018systematic}. With VR and computerized simulations, surgical tasks are infinitely customizable and repeatable, with obvious benefits on the wideness of the trained surgical skillset. Most importantly, VR allows to calculate quantitative and objective performance metrics \cite{moody2003objective} useful for progress monitorning and self-assessment of performance, which has been demonstrated to ameliorate the learning phase and improve skill learning \cite{macdonald2003self} \cite{Bric2016}.    
The validity of VR setups and their efficacy in transferring skills to the operating room was evidenced by Haque \textit{et al.} \cite{Haque2006}, while Aghazadeh \textit{et al.} \cite{Aghazadeh2016} demonstrated a statistical correlation between robotic performance in a simulated environment and robotic clinical performance.
Hung \textit{et al.} \cite{hung2013comparative} confirmed the feasibility of animal models, cadavers, dry-lab phantoms and VR simulators as effective training tools with a comparative study of their construct validity.

Recently, the versatility of virtual simulators has been exploited for the introduction of assistance into surgical robotics training curricola, with the aim of improving performances and enhancing the learning of psychomotor skills with visual and visuo-haptic cues \cite{hamza2019survey}. Robotic assistance is proven correlated to psychomotor learning in the context of human-robot collaboration \cite{heuer2015robot}, as it has been demonstrated that visual guidance and haptic feedback play a significant role in surgical error reduction \cite{caccianiga2021multi}.
The role of such assistive strategies in a real \textit{in-vivo} surgical scenario is still uncertain and must be assessed through an extensive clinical trial \cite{Fosch2022}. The vast majority of assisted training protocols implementing \ha consists of \textit{ad-hoc} systems \cite{Lin2014} \cite{shahbazi2013dual} \cite{hong2021simulation}, which are limited both in terms of tasks implemented and in terms of evaluation protocols. Indeed, few studies \cite{Enayati2018} \cite{mariani2018design} have evaluated the trainee's performance on multiple diverse tasks and over the course of multiple training days, and the role of force-based haptic assistance on skill retention and skill transfer is still unclear.

This work proposes an exploratory assessment of the role of haptic assistance in the context of surgical training and, specifically, to the transferrability of skills with a multi-day experimental protocol articulated in two phases, designed in order to highlight the difference in the transfer and retention of skills between a control group and an assisted group. 

The novelty of this work consists in the evaluation of how assisting surgical robotics training affects the surgical performance even after the training phase, when assistance is absent.

\section{Materials and methods}

\subsection{Haptic Assistance Methods for Surgical Training}


In the context of \ha, assistance is delivered through forces applied to the manipulators at the surgical console, which re-directs the motion of the surgeon's hands in case of improper or unsafe maneuvers. The magnitude and direction of the force are computed from the position and orientation of the \ac{ee}  relative to the operatory space and the position of objects in the scene, in a feedback fashion. 

Most of the assistance strategies implemented here will use the distance from the virtual \ac{ee} tip to the target or obstacle as the primary metric for determining the intensity of the feedback force or torque. Different surgical tasks and situations require a level of control over how the distance is taken into account, and for this reason a sigmoidal mapping function is employed for the normalization of the linear or angular error $E$ into a suitable interval. Specifically, such mapping is formulated as:
\begin{equation}
  f_{map}() = \frac{1}{1+e^{5\delta w(E-t-h)}}
  \label{eq:sigmoidalmap}
\end{equation}
with $\delta = +1$ for guidance towards targets and $\delta = -1$ for obstacle avoidance. Here:
\begin{itemize}
  \item $t$ is the assistance \textit{threshold}, hence the value at which the sigmoid starts to significantly increase from zero [mm]
  \item $h$ is the distance from the threshold at which $\frac{1}{2}$ of the maximum force is provided [mm]
  \item $w$ controls the \textit{width} of the linear region, hence the steepness of the curve [mm$^{-1}$]
\end{itemize}



The assistance forces \textbf{\textit{F}} and torques \textbf{\textit{T}} that have been implemented are the combination of an elastic component, proportional to the error mapped with the sigmoidal function, and a viscous component proportional to its rate of change, which damps the possible oscillatory instabilities. The optimal visco-elastic balance is heavily task-dependent and surgeon-dependent, and for this reason it is tuned by manually setting the values of $K_F$, $\eta_F$, $K_T$ and $\eta_T$ in the equations
\begin{equation}
  \vect{F} = K_F\cdot\vect{F}_{elastic} + \eta_F\cdot\vect{F}_{viscous} 
  \label{eq:forcevfbalance}
\end{equation}
\begin{equation}
  \vect{T} = K_T\cdot\vect{T}_{elastic} + \eta_T\cdot\vect{T}_{viscous} 
  \label{eq:torquevfbalance}
\end{equation}    
which will ultimately result in the haptic outputs provided to the actuators. 
For each of the proposed assistance strategies proposed below, the following actors are defined:
\begin{itemize}
    \item EE: End-Effector, the tip of the virtual surgical instrument, the position and orientation of which are controlled by the user at the surgical console. 
    \item Target: Element in the scene that the surgeon is asked to reach for, follow, or stay close to. In relation to specific task the target can be represented by a 1D trajectory, 2D surface or a 3D object.
    \item Obstacle: Element in the scene that the surgeon should avoid, ideally staying as far as possible from it. 
\end{itemize}

\subsubsection{Trajectory Guidance} 
\begin{figure*}
  \centering
      \includegraphics[width=\linewidth]{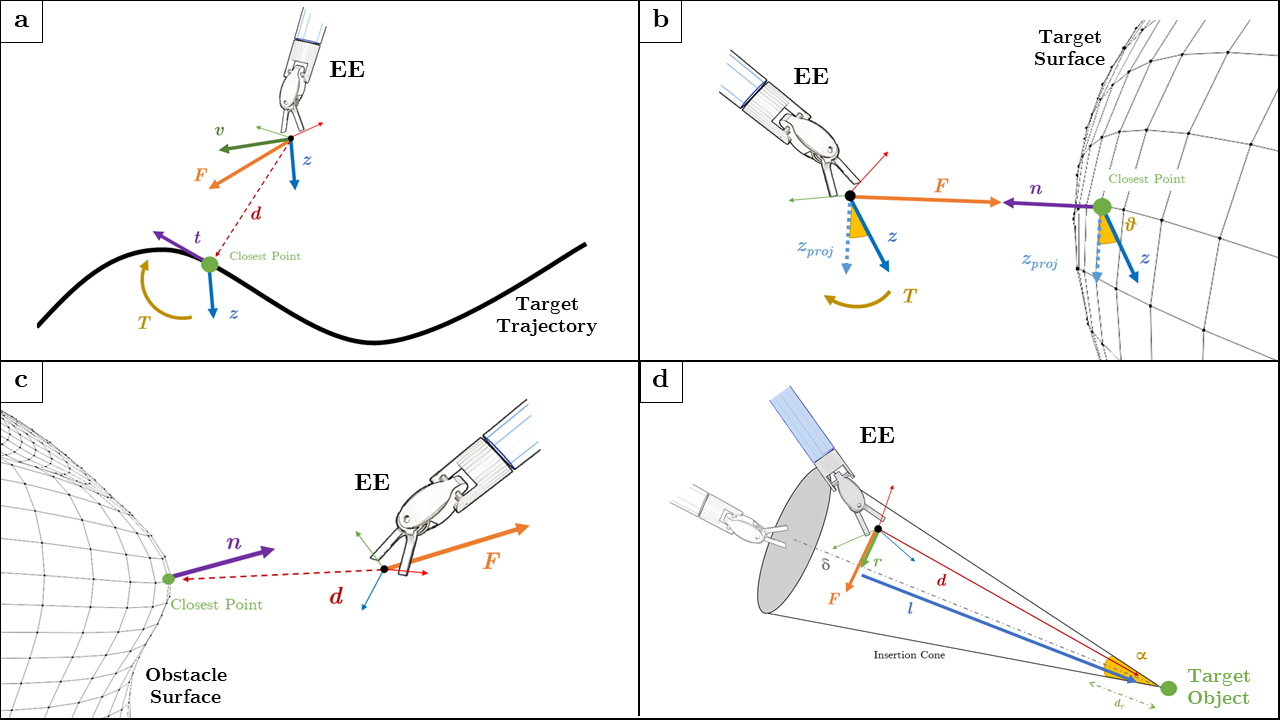}
      \caption{Vectors involved in the computation of the haptic assistance algorithms. \textbf{a.} Trajectory Guidance; \textbf{b.} Obstacle Avoidance; \textbf{c.} Surface Guidance; \textbf{d.} Insertion Guidance.}
      \label{fig:vfs_scheme_detailes}
\end{figure*}

Given a reference trajectory - planned in the pre-operative phase - this assistance steers the user's hands in order to align the robotic \ac{ee}  with the closest point on the trajectory. The feedback force attracts the \ac{ee}  towards the reference, while the torque rotates it so that it's aligned with the tangent vector at the closest point. \newline
Fig. \ref{fig:vfs_scheme_detailes}.a illustrates the vectors involved in the computation of the force assistance; specifically, contributions in Eq. \ref{eq:forcevfbalance} expanded as:
\begin{equation}
    \vect{F}_{elastic} = f_{map}(\|\vect{d}\|)\cdot\frac{\vect{d}}{\vect{d}}
\end{equation} 
\begin{equation}
    \vect{F}_{viscous} = 
    \begin{cases} 
        \vect{d}, & \textit{if         } \vect{v}\cdot\vect{d}<0 \\
         \text{rotate}(\vect{v},\theta,\vect{r}), & \textit{otherwise}
    \end{cases}
\end{equation} 

Here, $\vect{d}$ is the distance vector from the surgical instrument to the closest point in the trajectory, $\vect{v}$ is the velocity of the surgical instrument, the \textit{rotate}$(\cdot)$ function applies a rotational transform to rotate $\vect{v}$ of an angle  $\theta$ around axis $\vect{r}$, respectively:
\begin{align*}
    \theta = (1+\vect{v}\cdot\vect{d})\cdot\frac{\pi}{2}
    & & 
    \vect{r} = \vect{v}\times\vect{d}
\end{align*}
This implementation is adapted from \cite{Fan2022}. Similarly, contributions to the torque (Eq. \ref{eq:torquevfbalance}) are expanded as: 
\begin{equation}
    \vect{T}_{elastic} = \arccos (\vect{z}\cdot\vect{t} ) \cdot \vect{z} \times \vect{t}
    \label{eq:telastic}
\end{equation}
\begin{equation}
    \vect{T}_{viscous} = \frac{d}{dt} \left[ \arccos (\vect{z}\cdot\vect{t} ) \right] \cdot \vect{z} \times \vect{t}
    \label{eq:tviscous}
\end{equation}
In Eq. \ref{eq:telastic} and Eq. \ref{eq:tviscous}, the angle and axis of rotation which will achieve this alignment are $\arccos(\vect{z}\cdot\vect{t})$ and $\vect{z} \times \vect{t}$, respectively.

\subsubsection{Obstacle Avoidance} With this assistance algorithm, the distance from an obstacle object to the \ac{ee} and its rate of change are used for computing the force feedback, avoiding collisions of the \ac{ee} and the obstacle.\newline
The closest vertex point to the \ee is identified and, with $\vect{d}$ as the normalized distance vector going from such point to the \ee and $\vect{n}$ as the normal vector of the object mesh at that point (a graphic representation is depicted in Fig. \ref{fig:vfs_scheme_detailes}.b), the force is again computed as the sum of an elastic and a viscous component:
\begin{equation}
    \vect{F}_{elastic} = f_{map}(\|\vect{d}\|) \cdot \frac{\vect{n}}{\|\vect{n}\|} 
    \label{eq:obstacleavoidancefe}
\end{equation}
\begin{equation}
    \vect{F}_{viscous} = \frac{d}{dt} \left( f_{map}(\|\vect{d}\|)\right)\cdot \frac{\vect{n}}{\|\vect{n}\|} 
    \label{eq:obstacleavoidancefv}
\end{equation}
$f_{map}(\cdot)$ is the error mapping function with $\delta=-1$

\subsubsection{Surface Guidance} A target surface mesh is used here as a reference for guidance: this implementation computes a force and a torque that attracts and align the surgical \ee toward the surface itself. 
The distance vector to the closest point belonging to a reference surface determines the force feedback, while the torque is calculated from the angular error.
For the force computation, the distance vector $\vect{d}$ goes from the \ac{ee} to the closest point of the surface mesh, and $\vect{n}$ is the normal vector of the surface at that point. The elastic and viscous contributions to the force are:

\begin{equation}
    \vect{F}_{elastic} = f_{map}(\|\vect{d}\|) \cdot \frac{\vect{-n}}{\|\vect{n}\|} 
    \label{eq:surfaceguifancefe}
\end{equation}
\begin{equation}
    \vect{F}_{viscous} = \frac{d}{dt} \left[ f_{map}(\|\vect{d}\|)\right]\cdot \frac{\vect{-n}}{\|\vect{n}\|} 
    \label{eq:surfaceguifancefv}
\end{equation}

To compute the torque, one considers the relative orientation of the $\vect{z}$-axis of the \ee reference frame and the normal vector $\vect{n}$ at the closest point on the surface. The torque generated on the manipulator will aim at aligning the $\vect{z}$-axis with its projection on the tangent plane of the surface, defined by the normal vector $\vect{n}$ at the closest point. The vector projection on the tangent plane is the difference between the vector itself and its projection on the normal vector $\vect{n}$, obtained from the dot product operator:
\begin{equation}
    \vect{z}_{proj} = \vect{z} - (\vect{z}\cdot\vect{n})\vect{n}
\end{equation} 

From this, the torque is again the sum of an elastic component dependent on the angle $\theta$ between $\vect{z}$ and $\vect{z}_{proj}$ (specifically $\theta = \arccos(\vect{z}\cdot\vect{z}_{proj})$) , and a viscous one proportional to the angle's derivative in time. The formula for the torque is
\begin{equation}
    \vect{T}_{elastic} = f_{map}(\theta) \cdot \frac{\vect{z} \times \vect{z}_{proj}}{\|\vect{z} \times \vect{z}_{proj}\|}
    \label{eq:}
\end{equation}
\begin{equation}
    \vect{T}_{viscous} = \frac{d}{dt} \left( f_{map}(\theta)\right)\cdot \frac{\vect{z} \times \vect{z}_{proj}}{\|\vect{z} \times \vect{z}_{proj}\|}
    \label{eq:}
\end{equation}

where $\vect{z} \times \vect{z}_{proj}$ is the axis of rotation.

\subsubsection{Insertion Guidance} This assistance method aids the surgical insertion of the \ac{ee}  towards a target point, maintaining the path of the \ee stable inside an insertion cone. A similar implementation of this \ha algorithm, from which this algorithm is inspired, was proposed by Bettini \textit{et al.} \cite{Bettini2004}.
From a starting pose and a target object one defines an \textit{insertion cone} as a safe space for the \ee to move while approaching the target object. An insertion angle $\alpha$ determines the conical aperture and, therefore, how narrow the safe space is as the \ee gets closer to the target. The vector $\bm{\delta}$ goes from the initial \ee position to the target. 
\begin{align*}
    \vect{l} =  \left(\vect{d}\cdot \frac{\bm{\delta}}{\|\bm{\delta}\|} \right) \frac{\bm{\delta}}{\|\bm{\delta}\|}
 &  & 
    \vect{r} = \vect{d} - \vect{l}
\end{align*}

Fig. \ref{fig:vfs_scheme_detailes}.d shows a representative example.

For the insertion cone, fixed in space, the aperture $a$ relates the longitudinal and radial components through the angle $\alpha$ (expressed in degrees) as follows:
\begin{align*}
    a = \tan\left(\alpha\cdot\frac{\pi}{180}\right) &  & 
    \|\vect{r}\| = \|\vect{l}\|\cdot a
\end{align*}

The computed feedback force aims at maintaining the \ee position inside the safe insertion cone, therefore the threshold in the error mapping function $f_{map}(\cdot)$ is set to $t = \|\vect{r}\|$, the value of which varies over time and depends on the longitudinal coordinate $\vect{l}$. When the \ee is very close to the target, the threshold becomes very small and the force gets very close to the maximum value very rapidly, which causes instabilities. To solve this issue, a ``relax distance'' $d_r$ is introduced: when the \ee is close enough to the target, the force is relaxed and scaled down to half of its value to interfere less with the surgeon's motion so close to the target. Our clinical consultant suggested that $d_r$ should be 20\% of the height of the cone. The force feedback for the \textit{Insertion Guidance} assistance strategy is:
\begin{equation}
    \vect{F}_{elastic} = r(\|l\|)\left[\cdot f_{map}(\|\vect{a}\|)\right]\cdot \frac{\vect{a}}{\|\vect{a}\|}
    \label{eq:insertionguidance}
\end{equation}
\begin{equation}
    \vect{F}_{viscous} = r(\|l\|)\left[\cdot \frac{d}{dt} \left( f_{map}(\|\vect{a}\|)\right)\right]\cdot \frac{\vect{a}}{\|\vect{a}\|}
    \label{eq:insertionguidance}
\end{equation}

with

\begin{equation}
    r(\|l\|) = 
    \begin{cases}
        1 & \text{if } \|l\| > d_r \\
        0.5 & \text{if } \|l\| \leq d_r
    \end{cases}
\end{equation}

scaling the force down to its relaxed value depending on the insertion depth.
Here the feedback force has the same direction as the radial conical coordinate and a magnitude that is proportional both to the distance to the cone centerline and the distance to the target point. With this configuration, the \ac{ee} tip will be kept inside the reference insertion cone.

\begin{figure*}
  \centering
      \includegraphics[width=\linewidth]{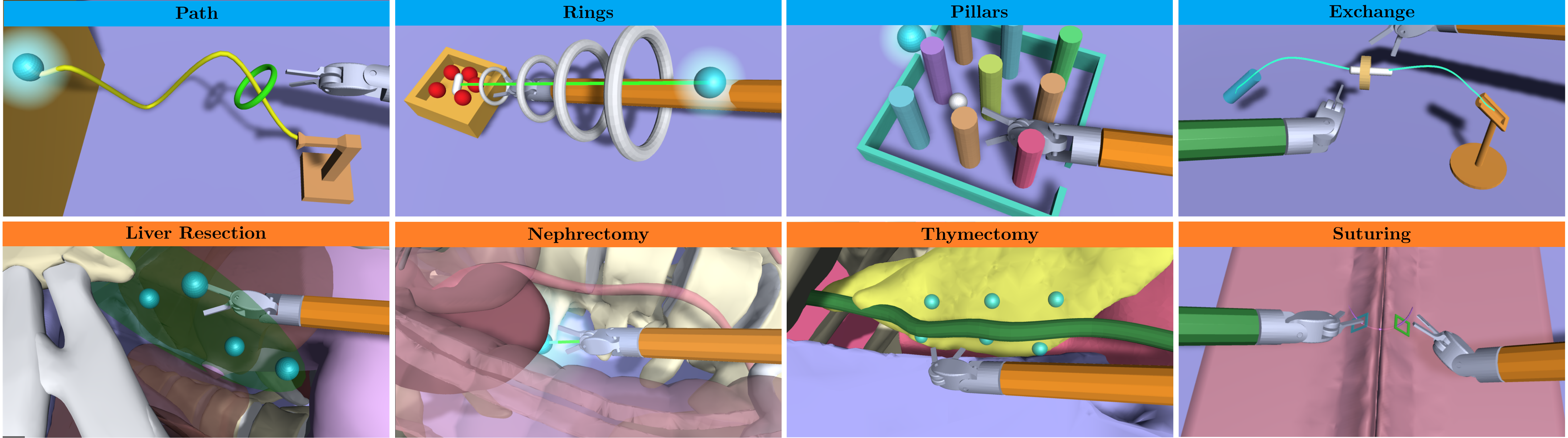}
      \caption{Snapshot of the simulated surgical tasks, with the respective denomination. Training tasks have blue headlines, while realistic evaluation tasks have orange headlines. Tasks on the same column share the same surgical skills required for their completion. \textit{Path} and \textit{Liver Resection} require articulate wrist motion and stability: the trainee is here asked to follow reference trajectories or to stay close to refence surfaces. They employ Trajectory Guidance and Surface Guidance, respectively. \textit{Rings} and \textit{Nephrectomy} survey the depth perception skills: the user should reach for targets in scenarios where depth-perception is impeded or non-trivial. Insertion Guidance is the assistance employed in these tasks. \textit{Pillars} and \textit{Thymectomy} are hand-eye coordination tasks: the trainee should perform gripping actions without colliding with nearby structures. assisted with Obstacle avoidance.\textit{Exchange} and \textit{Suturing}, both bi-manual tasks, challenge the capabilities in terms of instrument exchange, and require the user to correctly pass an object from one hand to the other. Both are assisted with Trajectory Guidance.}
      \label{fig:taskspanel}
\end{figure*}

\subsection{Surgical Simulator}

The simulator comprises eight surgical tasks, four of which (\textit{Path, Rings, Pillars} and \textit{Exchange}) are simplistic training tasks built with objects of simple geometry, while the remaining four (\textit{Liver Resection, Nephrectomy, Thymectomy} and \textit{Suturing}) emulate \textit{in-vivo} surgical procedures and are therefore more realistic. Fig. \ref{fig:taskspanel} collects snapshots of the tasks. All of these are constructed and set-up in order to be as challenging as possible in relation to a specific surgical skill. A set of fundamental pre-operative and intra-operative skills that any robotic surgeon should acquire during training was proposed in \cite{Smith2014}. 

\subsection{Surgical Training Hardware}
This research was conducted on a \textit{daVinci}\cright surgical robot integrated with the dVRK \cite{Kazanzidesf2014} framework. The \acp{mtm} of the dVRK \cright are in fact equipped with motors usually employed for the sake of homing and calibration; the \ha forces and torques are generated by energizing these motors according to the inverse dynamics model of the manipulators\cite{Fontanelli2017}. A ROS framework manages the communication between the teleoperation console of the dVRK and the virtual surgical scene, which is built upon the Unity\cright physics engine: therefore, the real \acp{psm} do not move during teleoperation.
The simulator also exploits the 3D viewing capability of the \ac{hrsv} installed on the teleoperation surgical console: two virtual cameras are positioned in the Unity\cright scene at a distance of 5.3mm, with their feed being sent separately on the left and right oculars at the console achieving a three-dimensional perception of the virtual environment.
The system runs at 30Hz.

\subsection{Experimental Protocol}
Two resident surgeons from the \textit{European Institute of Oncology} (IRCCS, Milan, Italy), both regularly performing RAMIS procedures with the \textit{daVinci}\cright robot, kindly dedicated their time in testing the surgical simulator in all its aspects, from the motion truthfulness to the complexity of the wrist articulation to the invasiveness and visco-elastic balance of the haptic force assistance. Their opinion and expertise were precious and insightful tools that guided the development towards a clinically validated robotic surgical simulator. 
Moreover, the most expert resident surgeon allowed to have his performance recorded when practicing with the simulator, which will be considered ``peak performance'' in the experimental analysis.

The effectiveness of this assisted training framework has been assessed with an experimental study where the performance of un-assisted subjects in a control group was compared to the one recorded from subjects to whom was provided haptic assistance. 8 novice subjects with little to no experience with surgical robots were recruited for the study. The subjects had diverse educational background (engineering, medicine, computer science) and had either never used a surgical robot or used it for less than 2 hours in total. Subjects were 25\% females and 75\% males, between 23 and 27 years of age, all right-handed and either had never teleoperated a surgical robot or did it less than 5 times. Assignation to the control or assisted group was random.
After a 15 minutes familiarization period in a ``Playground'' environment, the experimental phase was conducted as follows:
\begin{itemize}
    \item From Day 1 to Day 4 all the subjects executed the training tasks (\textit{Path}, \textit{Rings}, \textit{Pillars} and \textit{Exchange}) thrice a day: force-based haptic assistance was delivered only to the subjects in the assisted group. 
    \item On Day 5 and Day 6 no training was performed by neither of the groups. 
    \item On Day 7, both the assisted and the control group were asked to execute never-seen-before surgical evaluation tasks (\textit{Liver Resection}, \textit{Thymectomy}, \textit{Nephrectomy} and \textit{Suturing}): in this final phase, neither of the groups was assisted in the execution.
\end{itemize}
\noindent A quantitative estimation of surgical performance is obtained by combining metrics recorded in real-time during the execution of the task. The simulator logs these metrics autonomously detecting when the user initiates the execution and when the task is completed. The metrics are:
\begin{itemize}
      \item
      $D = \| \vect{p}_{EE} - \vect{p}_{targ/obs}\| $ 
      Distance Error, from the position of the \ee to position of the target or obstacle), measured in \textit{mm}
      \item
      $A = \text{acos}(\vect{z}_{EE}\cdot\vect{z}_{targ})$
      Angular Error, calculated from the z-axis of the \ee local reference frame and the tangent vector of the target , measured in \textit{rad}
      \item
      $F = f(D, \dot{D}, K, \eta) $
      Force Feedback Magnitude, calculated by the specific assistance algorithm as a function of the distance error $D$, its rate of change  $\dot{D}$ and the visco-elastic parameters, measured in \textit{N}
      \item
      $T = f(A, \dot{A}, K, \eta) $ 
      Torque Feedback Magnitude, calculated by the specific assistance algorithm as a function of the angular error $A$, its rate of change  $\dot{A}$ and the visco-elastic parameters, measured in \textit{Nm}
      \item
      $M$ the number of drops when exchanging an instrument, non-dimensional 
      \item
      $C$ the fraction of task time spent repositioning with respect to the total task time, non-dimensional 
\end{itemize}

These values are logged at each frame of the task execution and are then averaged once the task is complete. Metrics are combined with a weighted average to obtain a quantitative performance score: the weights are dependent on the task and the key surgical skills that such a task requires and have been determined with the help of our clinical consultants. Considering ``optimal execution'' the one achieved by the resident surgeon, the quantitative performance index $P$ for a task execution is the weighted average of the performance metrics normalized by the performance metric recorded from the expert surgeon.
\begin{equation}
    P = \frac{1}{10}\sum_{k=1}^{6}w_k \cdot \frac{X_k(subject)}{X_k(surgeon)}
\end{equation}
with $k \in \{D,A,F,T,M,C\}$ refers to each of the metrics included in this study. Weights are reported in Table \ref{tab:weights}.
\begin{center}
\begin{table}
\centering
    \label{tab:weights}
    \caption{Task-specific weights for the calculation of performance $P$}
    \begin{tabular}{||c||c|c|c|c|c|c||}
        \hline
        \textbf{Task} & $w_{\hat{D}}$ & $w_{\hat{A}}$ & $w_{\hat{F}}$ & $w_{\hat{T}}$ & $w_{\hat{M}}$ & $w_{\hat{C}}$ \\
        \hline\hline
        \textit{Path}            & 3 & 2 & 3 & 1 & 0 & 1 \\
        \hline
        \textit{Rings}           & 5 & 0 & 4 & 0 & 0 & 1 \\
        \hline
        \textit{Pillars}         & 5 & 0 & 4 & 0 & 0 & 1 \\
        \hline
        \textit{Exchange}        & 2 & 2 & 2 & 1 & 2 & 1 \\
        \hline
        \textit{Thymectomy}      & 5 & 0 & 4 & 0 & 0 & 1 \\
        \hline
        \textit{Nephrectomy}     & 5 & 0 & 4 & 0 & 0 & 1 \\
        \hline
        \textit{Liver Resection} & 3 & 2 & 3 & 1 & 0 & 1 \\
        \hline
        \textit{Suturing}        & 2 & 3 & 1 & 2 & 1 & 1 \\
        \hline
    \end{tabular}
\end{table} 
\end{center} 
\begin{figure*}[t]
  \centering
      \includegraphics[width=\linewidth]{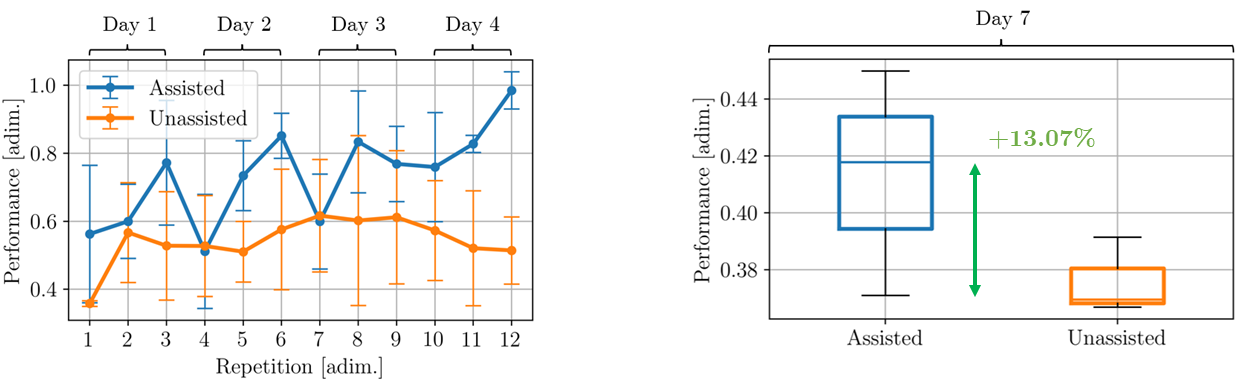}
      \caption{\textbf{Left: }Performance trend of assisted subjects (blue) and unassisted subjects (orange) of the \textit{Path} training task; \textbf{Right: }Boxplots of the average performance of assisted subjects (blue) and unassisted subjects (orange) of the \textit{Thymectomy} evaluation task}
      \label{fig:results_compact}
\end{figure*}
\section{Results}
Fig. \ref{fig:results_compact} shows the performance trends for one of the four training tasks, where the performance at each repetition is the average among the subjects in the assisted or control group. The trends are increasing both for the assisted group and the unassisted group. Most significantly, the performance in the assisted group is consistently higher than the one in the control group. The performance trends shown for \textit{Path} are representative of the other three training tasks (\textit{Rings, Pillars} and \textit{Exchange}). The performance on one of the four validation tasks recorded on the last day of the experimental phase is also shown in Fig. \ref{fig:results_compact} with boxplots. The graph reports the distribution of performances collected from the 4 subjects executing 3 repetitions. Crucially, neither the subjects in the assisted group nor the ones in the control group were guided with \ha on these tasks, nevertheless the performance recorded from the assisted subjects is distributed on higher values for all the tasks. The performance distribution shown for \textit{Nephrectomy, Liver Resection} and \textit{Suturing} mimic the one reported in Figure \ref{fig:results_compact}, which is referred to \textit{Thymectomy}.

Quantitative results are obtained by comparing, for each task, the mean, standard deviation and median values of performance between the assisted and the control group. Given the data scarcity and their non-Gaussian distribution, the most meaningful conclusions will be drawn from the median values. Apart from \textit{Nephrectomy} showing a slightly reduced median performance on assisted subject ($-2.71\%$), all other tasks present an increase in both the mean and median performance, as high as $+21.54\%$ for \textit{Liver Resection} (\textit{Thymectomy} $+13.07\%$, \textit{Suturing} $+8.44\%$).

\section{Discussion}
The results suggest that \ha grants a performance improvement when executing surgical tasks, an aspect that may be most beneficial in terms of safety and invasiveness when translated in the real surgical context. Under this light, haptic assistance effectively acts as an error-correction strategy which, when applied in real-time, re-directs the \ac{ee}  towards safer spatial regions by acting on the master manipulators gripped by the surgeon.
Concerning the training experience and the associated learning curve, the available results do not show any significant difference when comparing the two groups, and the hypothesized benefits of \ha regarding this aspect remain to be verified.

The most interesting considerations may be drawn from the difference in performance on the never-seen-before unassisted evaluation tasks, in favor of the assisted subjects. Since for these tasks, which were purposely designed to resemble real surgical scenarios, no haptic assistance was provided to either of the groups, it can be concluded that the introduction of haptic assistance in the training phase could contribute to the skill transfer from training tasks to surgical tasks. This is arguably due to the integration of the haptic guidance into the visuo-haptic motor feedback loop that acts during teleoperation. As a consequence, the benefits of employing haptic assistance could arise after the training phase as well, when Haptic Assistance is not in use. However, these assertions would benefit from a larger experimental study. 

Other than the population size and scale of the experimental study, limitations of this work include a fixed parametrization of the visco-elastic balance of the assistive force, which might benefit some users and impede others. Extensions of this work might include adaptive force control algorithms that tailor the visco-elastic parameters in real time. Additionally, the longer term (months, years) effect of force assisted training protocols is still uninvestigated.

\section{Conclusions}
This work features the development of a haptic-enhanced VR surgical simulator integrated with a surgical robot and an experimental study on the role of force-based haptic assistance employed as assistance strategies in the surgical training context. The results of the experimental study
suggest that employing \ha during the training phase of surgical practice leads to improved performance and might contribute to skill transfer toward real surgical scenarios where haptic assistance is absent.

\newpage
\bibliographystyle{unsrt}


\end{document}